%% file: lnbip.tex
\documentclass[lnbip]{svmultln}
\usepackage{makeidx}  % allows for indexgeneration
\usepackage{amsmath}
\usepackage{graphicx}
\usepackage[colorlinks=true, allcolors=blue]{hyperref}
\usepackage{booktabs}
\usepackage{pdflscape}
\usepackage{adjustbox}
\usepackage{array}
\usepackage{svg} 
\usepackage{multirow}
\usepackage{multicol}
\usepackage{tabularx}
\usepackage{arydshln}
\usepackage{amsmath}
\usepackage{bm}
\usepackage{graphicx}
\usepackage{subcaption}
\usepackage{enumitem}

\newcommand{\mypar}[1]{\smallskip\noindent\textbf{#1.}}

\begin{document}
\mainmatter              % start of the contribution
\title{On the Simplification of Neural Network Architectures for Predictive Process Monitoring}
\titlerunning{Simplification of Neural Network Architectures for PPM}  % abbreviated title (for running head)
\author{Amaan Ansari\inst{1,2} \and Lukas Kirchdorfer\inst{1,2} \and Raheleh Hadian\inst{1}}
\authorrunning{Amaan Ansari et al.}   % abbreviated author list (for running head)
\tocauthor{Amaan Ansari, Lukas Kirchdorfer, Raheleh Hadian}
\institute{SAP Signavio, Walldorf, Germany\\
% \email{first.last@sap.com}
\and
University of Mannheim,
Data and Web Science Group, Mannheim, Germany}

\maketitle              % typeset the title of the contribution
% \index{Ekeland, Ivar} % entries for the author index
% \index{Temam, Roger}  % of the whole volume
% \index{Dean, Jeffrey}

\begin{abstract}        % give a summary of your paper at least 70 and at most 150 words
Predictive Process Monitoring (PPM) aims to forecast the future behavior of ongoing process instances using historical event data, enabling proactive decision-making. While recent advances rely heavily on deep learning models such as LSTMs and Transformers, their high computational cost hinders practical adoption. Prior work has explored data reduction techniques and alternative feature encodings, but the effect of simplifying model architectures themselves remains underexplored. In this paper, we analyze how reducing model complexity---both in terms of parameter count and architectural depth---impacts predictive performance, using two established PPM approaches. Across five diverse event logs, we show that shrinking the Transformer model by 85\% results in only a 2–3\% drop in performance across various PPM tasks, while the LSTM proves slightly more sensitive, particularly for waiting time prediction. Overall, our findings suggest that substantial model simplification can preserve predictive accuracy, paving the way for more efficient and scalable PPM solutions.
%                         please supply keywords within your abstract
\end{abstract}
%

% input content here
% \input{text/content}

\input{Sections/01_Introduction}
\input{Sections/02_Related_Work}

\input{Sections/03_Methodology}

\input{Sections/04_Evaluation}

\input{Sections/05_Conclusion}

\bibliographystyle{splncs}
\bibliography{bibliography}
\end{document}

%% file: Sections/01_Introduction.tex
%Start 01_Introduction.tex
\section{Introduction}
\label{sec:intro}
% context
Predictive Process Monitoring (PPM) has become a central topic in Business Process Management (BPM), shifting the focus from retrospective analysis to forecasting future aspects of ongoing process instances. By anticipating properties such as the next activity, the next resource, upcoming timestamps, remaining time, or the final outcome, PPM enables organizations to make proactive decisions, optimize resource allocation, and mitigate potential risks \cite{di2022predictive,rama2021deep,teinemaa2019outcome}.

Recent advances in PPM have been largely driven by deep learning, with Recurrent Neural Networks (RNNs) and in particular Long Short-Term Memory (LSTM) networks proving effective in modeling the sequential nature of event logs \cite{TaxVRD17,Camargo2019}. More recently, Transformer-based models, originally developed for Natural Language Processing (NLP), have been successfully adapted for PPM. Models such as ProcessTransformer \cite{Bukhsh21} and MTLFormer \cite{WangHMLWY23} leverage self-attention mechanisms to capture long-range dependencies in event sequences---an area where traditional RNNs often struggle.

% problem
Despite their predictive power, these deep learning models are computationally intensive, with large numbers of parameters and high training costs. This complexity presents a barrier to practical deployment, particularly in industrial settings where computational resources are limited. For instance, process mining vendors like SAP Signavio, Celonis, or Apromore serve a large number of diverse clients across many domains, each with unique processes. Delivering PPM capabilities at scale would require training and maintaining separate models for each process and customer, leading to high costs and limited scalability.

% contributions
In this paper, we investigate whether pruning of state-of-the-art deep learning architectures can deliver competitive performance while significantly reducing computational overhead. We focus on streamlined variants of the LSTM-based model proposed by Camargo et al. \cite{Camargo2019} and the Transformer-based MTLFormer introduced by Wang et al. \cite{WangHMLWY23}, evaluating their potential as lightweight alternatives for PPM in resource-constrained environments.
Through our experimental investigation, we present the following key findings:
\begin{itemize}[noitemsep,topsep=0pt]
    \item We found that our simplified Transformer models achieve prediction performance similar to the full MTLFormer architecture, being on par or only slightly worse on average across several PPM tasks, but with a significant reduction in model parameters of around 85\%. This suggests that substantial parameter reduction in Transformer models is possible without a proportional loss in performance for PPM tasks.
    \item In contrast, simplifying the LSTM architecture to a similar degree in terms of model parameters often leads to a more pronounced performance degradation, showing a performance drop of 3-13\% on average across the core PPM tasks.
    \item Besides these, our experiments revealed a few more interesting findings, such as the respective strengths and weaknesses of LSTM and Transformer models for the mentioned PPM tasks. %, or about the encoding of time features in Transformer models.
\end{itemize}

% remainder
\noindent
The remainder of this paper is structured as follows: \autoref{sec:rel_work} provides the necessary background on PPM and discusses related work. Then, \autoref{sec:methodology} introduces the required concepts and presents the employed architectures and the training procedure, followed by our evaluation in \autoref{sec:evaluation}. Finally, we conclude the work in \autoref{sec:conclusion}.

%% file: Sections/02_Related_Work.tex
\section{Background and Related Work}
\label{sec:rel_work}
In this section, we provide the necessary background on PPM and survey existing efforts to improve the computational efficiency of these predictive models.

\subsection{Predictive Process Monitoring}
PPM has emerged as a promising area within process mining, offering a wide range of business applications. Early research in this domain focused primarily on predicting process outcomes, particularly in terms of duration and successful completion. To this end, a variety of techniques have been proposed, e.g., based on Hidden Markov Models \cite{Pandey2011}, finite state machines \cite{Folino2013}, or stochastic Petri nets \cite{Rogge2013}. In addition, classical machine learning approaches---such as random forests and support vector machines---have been adapted to predict process outcomes using handcrafted features derived from event histories \cite{Conforti2013}.

With the advent of deep learning, and in particular LSTM networks, the focus shifted toward automated feature learning from large-scale, sequential event logs. Early work by Evermann et al. \cite{EvermannRF17} demonstrated the effectiveness of shallow LSTM models combined with embedding techniques for categorical variables in predicting the next event. Building on this, Tax et al. \cite{TaxVRD17} used a similar LSTM architecture with one-hot encodings to predict the next activity, its corresponding timestamp, the remaining time, and the full process suffix. Camargo et al. \cite{Camargo2019} extended this approach by combining multiple LSTM components to support both categorical and numerical features.

Beyond LSTM-based models, Transformer architectures have gained traction in PPM. Bukhsh et al. \cite{Bukhsh21} introduced the ProcessTransformer to jointly predict the next activity, the next timestamp, and the remaining time. Wang et al. \cite{WangHMLWY23} further advanced this line of work with the MTLFormer, a multi-task learning model that addresses all three prediction tasks within a unified framework. Graph Transformer architectures have also been explored, particularly for the task of remaining time prediction \cite{Amiri2024}.

More recent research has leveraged PPM models also in the context of process simulation, either as a component of the simulation model \cite{Camargo2019} or for the evaluation of the simulation model's quality \cite{ozdemir2025rethinking}.

% Additionally, Convolutional Neural Networks (CNNs) have been successfully applied to PPM tasks, as demonstrated in the work of Pasquadibisceglie et al. \cite{pasquadibisceglie2019using} and Weytjens et al. \cite{weytjens2020process}, showing that local temporal patterns in event logs can also be effectively captured using convolutional layers.

\subsection{Towards More Efficient PPM Models}
The deep learning models discussed above, while effective, often demand substantial computational resources and long training times, posing significant challenges for large-scale deployment in resource-constrained environments. Several studies in the PPM community have approached this issue from different perspectives. One line of work focuses on reducing the size of the training dataset. For example, Sani et al. \cite{Sani2023} propose a sampling strategy to select an informative subset of the data for model training. Pauwels et al. \cite{Pauwels2021} address the problem by introducing incremental learning techniques that update neural networks with new data, thereby avoiding full retraining.
Another strategy involves exploring alternative model architectures to improve efficiency. Weytjens and De Weerdt \cite{WEYTJENS2022109134} demonstrate that Convolutional Neural Networks (CNNs) can offer both faster training and improved accuracy compared to LSTMs. Additionally, modifications to sequence encoding techniques have been proposed to further enhance training efficiency. Specifically, Roider et al. \cite{Roider2024} propose trace encoding as an alternative to prefix sequence encoding.

While these approaches have shown promise in reducing training time and resource usage, none have explicitly investigated whether simplifying the architecture itself, by reducing the number of parameters and overall model size, can yield efficient PPM solutions without compromising predictive performance. This gap motivates our investigation into lightweight model variants for PPM.

%% file: Sections/03_Methodology.tex
%Start 03_Methodology.tex
\section{Methodology}
\label{sec:methodology}
This section outlines the methodology of our study. We start by introducing the key concepts and defining the various PPM tasks. We then describe the MTLFormer \cite{WangHMLWY23}, the LSTM model \cite{Camargo2019}, and our simplified variants of these architectures. Finally, we detail the training procedure.

\subsection{Preliminaries and Problem Definition}
We begin by defining core concepts such as event log, trace, and prefix before formally defining the PPM tasks.

\mypar{Event log, traces, prefixes}
We assume that process execution data is stored in an event log $\mathcal{E}$, defined as a finite set of traces. Each trace  $\sigma = \langle e_{1}, e_{2}, ..., e_{n}\rangle$ represents the ordered sequence of events for a single process instance (case). 
We define an event $e = (c, a, r, t_s, t_e)$ as a tuple consisting of a case ID $c$, an activity $a$, a role $r$, a start timestamp $t_s$, and an end timestamp $t_e$.
% Furthermore, for each element of the event tuple, we define a function that maps this event to this element. For instance, $\pi_A$ is a function that maps $e$ to an activity such that $\pi_A(e) = a$.
For each element \(l\in\{c,a,r,t_s,t_e\}\) of an event $e$, we define a projection function \(\pi_L\) such that \(\pi_L(e)\) returns that element (e.g. \(\pi_A(e)=a\)).
% Let $\pi_C$, $\pi_A$, $\pi_R$, $\pi_{T_s}$, and $\pi_{T_e}$ be functions that map an event $e = (c, a, t_s, t_e, r)$ to a case ID such that $\pi_C(e) = c$, to an activity such that $\pi_A(e) = a$, to a role such that $\pi_R(e) = r$, to a start timestamp such that $\pi_{T_s}(e) = t_s$, and to an end timestamp such that $\pi_{T_e}(e) = t_e$.
PPM is typically formulated as a classification or regression problem and begins with a feature extraction step. This involves generating prefixes of varying lengths from each completed trace to represent different execution stages.
Formally, for a given trace $\sigma$, we define its prefix of length $k \in [1, n-1]$ as $p^k(\sigma)=\langle e_{1}, ..., e_{k}\rangle$, representing a partial execution up to the $k$-th event of a trace with in total $n$ events. Each prefix is then encoded into a feature vector $\bm{x} \in \mathcal{X}$ and associated with one or more target values $y_t \in \mathcal{Y}$, each corresponding to a specific PPM task.

\mypar{Definition of PPM tasks}
We consider five common PPM tasks, which we formally introduce in the following.  Each task takes as input an event prefix $p^k(\sigma)$ of a trace $\sigma$.

\begin{enumerate}[noitemsep,topsep=0pt]
    \item \textbf{Next activity prediction}: Defined as $\Theta_a(p^k(\sigma)) = \pi_A(e_{k+1})$, predicting the activity of the next event.
    
    \item \textbf{Next role prediction}: Defined as $\Theta_r(p^k(\sigma)) = \pi_R(e_{k+1})$, predicting the role responsible for the next event.
    
    \item \textbf{Next event duration prediction}: Defined as $\Theta_d(p^k(\sigma)) = \pi_{T_e}(e_{k+1}) - \pi_{T_s}(e_{k+1})$, predicting the difference between the next event’s end and start timestamps.
    
    \item \textbf{Next waiting time prediction}: Defined as $\Theta_{wt}(p^k(\sigma)) = \pi_{T_s}(e_{k+1}) - \pi_{T_e}(e_k)$, predicting the difference between the next event’s start timestamp and the current event’s end timestamp.
    
    \item \textbf{Remaining time prediction}: Defined as $\Theta_{rt}(p^k(\sigma)) = \pi_{T_e}(e_n) - \pi_{T_e}(e_k)$, predicting the difference between the timestamp of the case’s last event and the current event’s end timestamp.
\end{enumerate}

%% before change the following code was used

% \mypar{Definition of PPM tasks}
% We consider five common PPM tasks, which we formally introduce in the following.

% \mypartwo{Next activity prediction}
% Given a prefix $p^k(\sigma)$ of a trace $\sigma$, next activity prediction can be defined as a function $\Theta_a$ that takes as input an event prefix $p^k(\sigma)$ and predicts the next activity such that $\Theta_a(p^k(\sigma)) = \pi_A(e_{k+1})$.

% \mypartwo{Next role prediction}
% Similarly to predicting the next activity, next role prediction can be defined as a function $\Theta_r$ that takes as input an event prefix $p^k(\sigma)$ and predicts the next role such that $\Theta_r(p^k(\sigma)) = \pi_R(e_{k+1})$.

% \mypartwo{Next event duration prediction}
% For a prefix $p^k(\sigma)$, next event duration prediction can be defined as a function $\Theta_{d}(p^k(\sigma)) = \pi_{T_e}(e_{k+1}) - \pi_{T_s}(e_{k+1})$
% to predict the difference between the next event's end and start timestamps.

% \mypartwo{Next waiting time prediction}
% For a prefix $p^k(\sigma)$, next waiting time prediction can be defined as a function $\Theta_{wt}(p^k(\sigma)) = \pi_{T_s}(e_{k+1}) - \pi_{T_e}(e_k)$
% to predict the difference between the next event's start timestamp and the current event's end timestamp.

% \mypartwo{Remaining time prediction}
% Given $p^k(\sigma)$, remaining time prediction can be defined as a function $\Theta_{rt}(p^k(\sigma)) = \pi_{T_e}(e_n) - \pi_{T_e}(e_k)$
% to predict the difference between the timestamp of the case's last event $e_n$ and the timestamp of the current event $e_k$.

%% ---

\subsection{Architectures}

We evaluate five architectures spanning both Transformer- and LSTM-based designs: the original MTLFormer \cite{WangHMLWY23}, our light variant of the MTLFormer, our model with a single transformer encoder (Transformer\textsubscript{simple}), instead of multiple, the original LSTM model proposed by Camargo et al. \cite{Camargo2019}, and our light variant of this.

\begin{figure}
    \centering
    \includegraphics[width=0.5\linewidth]{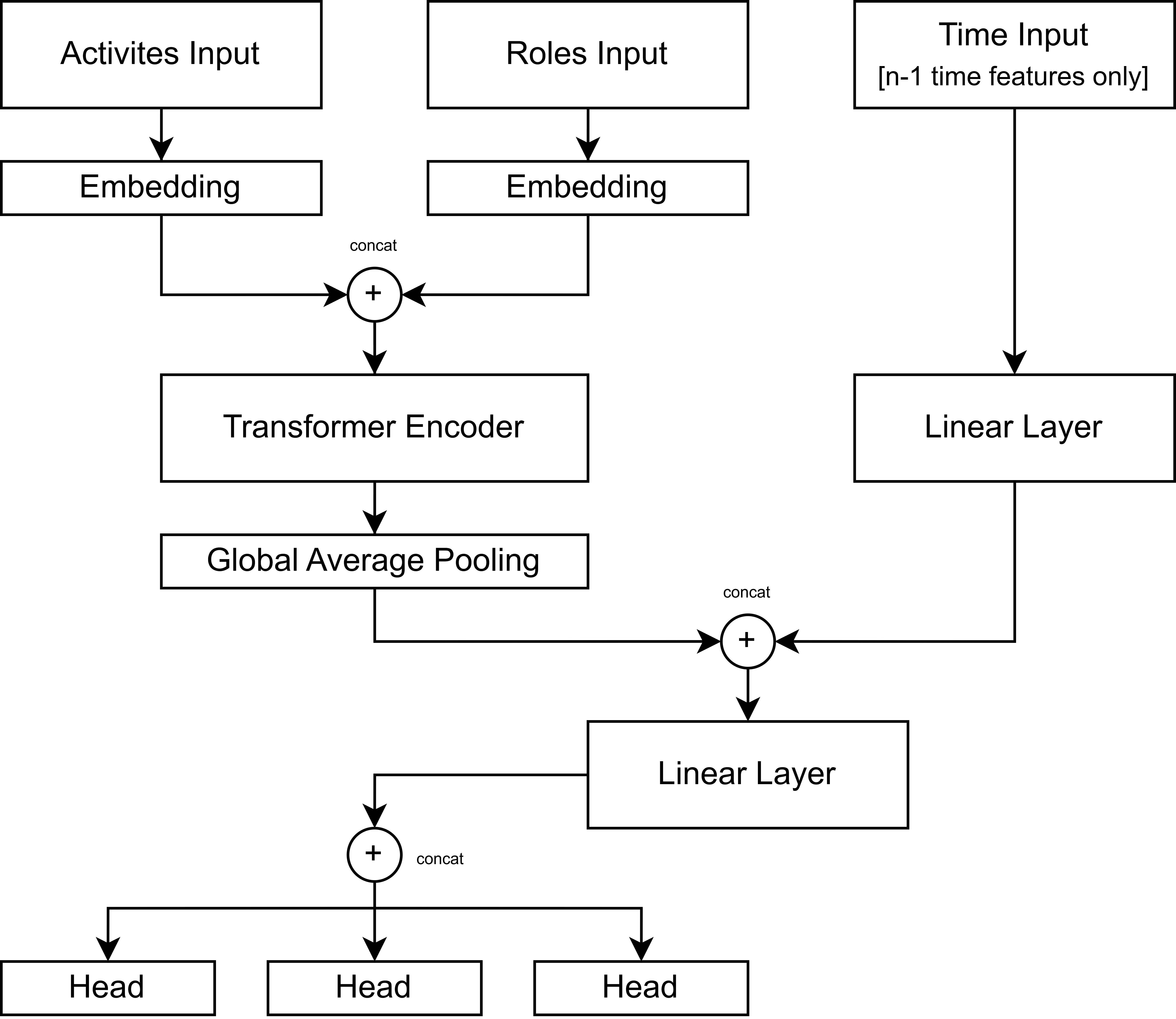}
    \caption{Our Transformer\textsubscript{simple} architecture}
    \label{fig:Architecture Transformer Simple}
\end{figure}

\mypar{Transformer Models}
We begin by describing the three Transformer models:
\begin{itemize}[noitemsep,topsep=0pt]
  \item  \textit{MTLFormer} uses five parallel Transformer streams, two of which are dedicated to activity labels, two to role labels, and one to temporal context. We denote these streams as \emph{activity stream 1}, \emph{activity stream 2}, \emph{role stream 1}, \emph{role stream 2}, and the \emph{temporal stream}. Each of these streams ingests its corresponding feature set (activity, role, or temporal), passes it through a Transformer encoder, and then applies average pooling to yield a fixed-length embedding. A Transformer encoder stacks multiple layers, each performing multi-head self-attention (MHA). The embeddings from activity stream 1, role stream 1, and the temporal stream are concatenated and projected. This projection is then concatenated with the embeddings from activity stream 2 and role stream 2. The concatenated representation is then fed into three deep multi-layer perceptrons (MLPs) acting as prediction heads, which forecast the next activity, the next role, and time-related values. Originally, the MTLFormer \cite{WangHMLWY23} predicts solely the next activity, the next timestamp, and the remaining time. However, we have extended both the inputs and the output heads to incorporate role prediction and waiting-time estimation, without altering the core architecture.
  \item \textit{MTLFormer\textsubscript{light}} consists of a backbone and three prediction heads. It preserves the original five‐stream Transformer backbone architecture of the MTLFormer while simplifying each prediction head to a single linear layer, instead of using MLPs. In addition, the number of parameters is reduced by shrinking the hyperparameters of the backbone. These include, but not limited to, the number of heads, the input embedding size and the feed-forward dimensions of the MHA. This modification is designed to reduce both parameter count and inference time without altering the backbone architecture. 
  %By keeping the five‐stream backbone architecture intact while cutting down parameter count, MTLFormer\textsubscript{light} lets us isolate the benefit of its multi‑stream design versus a simpler architecture, independent of model size.
  \item \textit{Transformer\textsubscript{simple}} (see~\autoref{fig:Architecture Transformer Simple}) simplifies MTLFormer\textsubscript{light} even further by removing its five parallel encoder streams and processing activity and role tokens within a single Transformer encoder. This yields a single Transformer encoder as the backbone, with three prediction heads, each of which consists of a single linear layer. The tokens are embedded, concatenated channel-wise, and passed through the Transformer encoder. The resulting feature maps are average-pooled and combined with a linear projection of the temporal features from the final input position. A shallow linear layer with ReLU activation function then feeds three single-layer heads for next activity, next role, and time predictions. This single-encoder design offers a substantially leaner architecture, enabling a direct comparison with MTLFormer\textsubscript{light}.

\end{itemize}

\mypar{LSTM Models}
Next, we describe the two LSTM models:
\begin{itemize}[noitemsep,topsep=0pt]
  \item \textit{LSTM}---with which we refer to the model proposed by Camargo et al. \cite{Camargo2019}---embeds activity, role, and temporal features, concatenates them, and feeds the sequence into a shared LSTM layer with batch-normalisation and dropout. This shared LSTM layer acts as the backbone. Afterwards, the last hidden state of the shared LSTM layer is routed into three parallel, task-specific heads, one each for activity, role, and time. Each head consists of a single LSTM layer and a small MLP.
  \item \textit{LSTM\textsubscript{light}}  retains the shared LSTM layer as backbone, but changes the prediction heads. Instead of using a single LSTM layer and a small MLP, it discards both the LSTM layer and the MLP altogether, using only a single linear layer inside each prediction head. All outputs are predicted directly from the shared LSTM layer backbone via a single linear projection per head, collapsing depth to reduce computation and parameters.
\end{itemize}

\subsection{Training Procedure}

\mypar{Data Preprocessing} We employ distinct pipelines for the LSTM and Transformer architectures. In each pipeline, we begin by computing three (normalized) temporal features: duration (length of each activity), waiting time (interval between activities), and remaining time until case completion.

We prepend each case with two special events, “start” and “end”, before we compute and normalize its time-based features. The "start" event marks the first activity from which the Transformer begins predicting what comes next, and the "end" event tells the model that the case has finished.  Finally, similar to the MTLFormer \cite{WangHMLWY23}, all models are trained on prefixes of each case, ranging from length 1 to $n-1$, where $n$ denotes the total number of events in the case. To build the training data, we extract every possible prefix of the chosen feature sequence: preceding activities for activity prediction, preceding roles for role prediction, or preceding time features for timestamp prediction. In addition, we pad each prefix to the length of the longest case so it fits the Transformer’s input size. Each padded prefix becomes the model input, and the very next feature in the sequence (the following activity, role, or time feature) serves as the target.

For the LSTM models, we adopt the preprocessing from Camargo et al. \cite{Camargo2019}, which closely mirrors the Transformer pipeline but replaces prefix padding with n-gram construction. The only modification we introduce is to unify the normalization step: both pipelines now use the same z-score parameters. In practice, this means each numerical feature is scaled using the identical mean and standard deviation across both the LSTM and Transformer approaches. 

\mypar{Loss Functions} During training, we compute a separate loss for each head, that is, one for next-activity prediction, one for next-role prediction, and one for the three time features, resulting in three concurrent losses. Although this third head outputs three continuous time-related values rather than a single scalar, it still counts as one head, and we compute MSE jointly over all three predictions. Cross-entropy loss is applied to the categorical tasks (activity and role), while MSE loss governs the continuous time-feature predictions. To balance these multitask objectives, we employ uncertainty weighting \cite{kendall2018multi} to combine them into a single scalar training loss.

\mypar{Hyperparameter Settings} We performed separate grid searches optimized for Transformer-based and LSTM-based models.

For the Transformer‐based models we explored embedding dimensions of 16 or 32 (ensuring divisibility by the number of heads), one, two or four attention heads, feed-forward dimensions of 32, 64 or 128, a fixed block dropout of 0.1, learning rates of 3e-4 or 6e-4, batch sizes of 8, 16 or 32, and one, two or four transformer encoder layers. 

In contrast, for the LSTM models we varied learning rates across 5e-4, 1e-3, 5e-3, 3e-4, and 6e-4, batch sizes between 8, 16, 32 and 64, n-gram sizes of 5, 10, and 15, hidden layer sizes of 50, 25, and 10, a fixed tanh activation, thus ensuring a thorough hyperparameter exploration for both architectures.

%% file: Sections/04_Evaluation.tex
%Start 04_Evaluation.tex
\section{Evaluation}
\label{sec:evaluation}
This section outlines the experiments conducted to evaluate the performance of the different Transformer and LSTM models.
In the remainder, \autoref{sec:setup} describes the experimental setup, followed by the results in \autoref{sec:results}. The implementation can be found in our repository\footnote{\url{https://github.com/amaanansari/simplified-ppm-models.git}}.

\subsection{Experimental Setup}
\label{sec:setup}

\mypar{Datasets}
Our evaluation is based on 5 event logs\footnote{Datasets available at \url{https://zenodo.org/records/5734443}}, spanning domains such as financial services, procurement, and manufacturing. As detailed in \autoref{tab:datasets}, these logs differ significantly in key properties such as the number of traces, events, activities, and resources. Crucially, all datasets contain both start and end timestamps for each event, allowing us to distinguish between the prediction tasks of next event time and next waiting time. 
Each dataset is sorted chronologically and then split horizontally into training, validation, and test sets, comprising 70\%, 10\%, and 20\% of the cases, respectively.

\input{Tables/datasets}

\mypar{Evaluation Metrics }
To assess predictive performance, we employ F\textsubscript{1}-score for the categorical tasks of next-activity and next-role prediction. For the continuous time-related tasks (waiting time, activity duration, and remaining time), we report the mean absolute error (MAE) in days. 

\mypar{Model Selection} 
% Due to our large hyperparameter grid, we use a composite score to rank each candidate model \(M\in\{\text{MTLFormer}_{\text{light}},\,\text{Transformer}_{\text{simple}}\}\) via its parameter‐count ratio and (weighted) relative validation loss difference to the best MTLFormer model. The best MTLFormer model was selected by the lowest validation loss. Concretely, let
% \[
% p_M \;=\;\frac{\#\text{params}(M)}{\#\text{params}(\text{MTLFormer})},
% \qquad
% \ell_M \;=\;\frac{\mathcal{L}(M)-\mathcal{L}(\text{MTLFormer})}
% {\mathcal{L}(\text{MTLFormer})}.
% \]
% Then define the overall score $S_M \;=\; p_M \;+\; 2\,\ell_M$, where the loss term is doubled to emphasize performance degradation relative to parameter savings. We then select the model \(M\) with the smallest \(S_M\). We apply the same principle for the LSTM models.
Let \(\mathcal{C}\) be the set of all candidate models produced by our hyperparameter grid, where each \(M\in\mathcal{C}\) denotes a specific architecture–hyperparameter configuration.
Let 
$M^\star \;=\; \arg\min_{M\in\mathcal{C}} \,\mathcal{L}(M)$
be the model with the lowest validation loss.
For any \(M\in\mathcal{C}\), define
\[
p_M \;=\; \frac{\#\mathrm{params}(M)}{\#\mathrm{params}(M^\star)},
\qquad
\ell_M \;=\; \frac{\mathcal{L}(M)-\mathcal{L}(M^\star)}{\mathcal{L}(M^\star)}.
\]
Here, \(\#\mathrm{params}(M)\) denotes the number of \emph{trainable} parameters of \(M\), and \(\mathcal{L}(M)\) is the validation objective (the same loss used during training) computed on the validation split.
We define the composite score
$S_M \;=\; p_M \;+\; \lambda\,\ell_M$,
and select the model with the smallest score, \(\arg\min_{M\in\mathcal{C}} S_M\).
We set \(\lambda=2\) unless stated otherwise to emphasize validation performance relative to parameter savings. We apply this scoring procedure independently within each model type (e.g., MTLFormer\textsubscript{light}, Transformer\textsubscript{simple}, LSTM, LSTM\textsubscript{light}) and for each dataset, with one model selected per type and dataset.

\subsection{Results}
\label{sec:results}

\input{Tables/Transformer_results_LK}

\input{Tables/LSTM_results_LK}

% \mypar{Overall Results}
Across five diverse event logs, we observe for both Transformer and LSTM models that their respective simplified variants achieve an enormous reduction in parameter count, on average over 80\% fewer parameters, while only incurring minimal losses in predictive performance.

Looking at the Transformer-based architectures in more detail (see \autoref{tab:results_transformer}), across five diverse event logs, MTLFormer$_\text{light}$ reduces the number of parameters by 85\% (from 136,412 to 19,823) compared to the full MTLFormer, while experiencing a mere 1.4\% decrease in the next-activity F\textsubscript{1} score (NAP, 0.70 → 0.69 F\textsubscript{1}) and a 2.8\% decrease in the next-role F\textsubscript{1} score (NRP, 0.71 → 0.69 F\textsubscript{1}). Mean absolute errors for next wait time prediction (NWTP) increased by just 6.3\%; next duration prediction (NDP) remained unchanged; and remaining time prediction (RTP) rose by only 0.7\%. 
Meanwhile, Transformer$_\text{simple}$, which not only reduces the parameter count (136,412 → 20,264) but also simplifies the overall architecture, largely matches MTLFormer$_\text{light}$ on the time‐related prediction tasks (NWTP MAE +7.0\%, NDP unchanged, RTP MAE +1.7\%). 
% Meanwhile, Transformer$_\text{simple}$, despite featuring an even more simplified architecture, achieves a comparable size reduction in terms of parameter count (136,412 → 20,264) and largely matches MTLFormer$_\text{light}$ on the time‐related prediction tasks (NWTP MAE +7.0\%, NDP unchanged, RTP MAE +1.7\%). 
However, its next‑activity prediction is, on average, about 3 percentage points lower (0.66 vs. 0.69 F\textsubscript{1}), a drop driven primarily by the larger gap on BPI12, the biggest dataset in our evaluation.  
Overall, these results suggest that a single transformer encoder can suffice for time‐related predictions, though the more elaborate MTLFormer$_\text{light}$ still holds a slight edge in next‐activity accuracy.

Looking at the LSTM in \autoref{tab:results_lstm}, the LSTM$_\text{light}$ model reduces its larger variant by 77\% (75,876 → 17,193 parameters), while incurring only minimal losses for the categorical prediction tasks. More specifically, we see a 2.9\% drop in both next-activity (0.70 → 0.68 F\textsubscript{1}) and next-role (0.70 → 0.68 F\textsubscript{1}) prediction. However, time-related errors are more affected: the next wait time MAE increases by 13\%, the remaining time MAE by 3\%, and the next duration MAE remains unchanged.

Compared to Transformer$_\text{simple}$, the LSTM$_\text{light}$ uses 15\% fewer parameters (17,193 vs 20,264) and slightly improves next-activity prediction (0.68 vs 0.66 F\textsubscript{1}), while matching next-role accuracy (0.68 F\textsubscript{1}). However, it shows notably worse timing performance, most strikingly in remaining time prediction, with a 26.8\% higher error (20.80 vs 16.40 MAE). Thus, while LSTM$_\text{light}$ is slightly more compact, Transformer$_\text{simple}$ offers superior temporal accuracy.

% When we compare the LSTM with our Transformer$_\text{simple}$, we see that the LSTM$_\text{light}$ is more compact, using 15\% fewer parameters (17,193 vs 20,264). It outperforms Transformer$_\text{simple}$ by 3\% in next-activity prediction (0.66 vs 0.68 F\textsubscript{1}), while matching it in next-role prediction (0.68 vs 0.68 F\textsubscript{1}). However, it yields modestly higher timing errors: next wait time increases by 2\% (1.53 → 1.56 MAE), next duration increases by 12.5\% (0.08 → 0.09 MAE) and remaining time increases by 26.8\% (16.40 → 20.80 MAE). Transformer$_\text{simple}$ thus offers superior temporal accuracy, albeit at a slightly larger footprint. Conversely, LSTM$_\text{light}$ trades precise time forecasts for an even smaller model and marginally superior activity F\textsubscript{1}. Notwithstanding the comparatively substantial time errors observed in relation to the Transformer$_\text{simple}$, we conclude that the Transformer$_\text{simple}$ demonstrates a slightly better overall performance in comparison to the LSTM$_\text{light}$.

\begin{figure}[htbp]
  \centering
  \begin{subfigure}[b]{0.45\textwidth}
    \includegraphics[width=\linewidth]{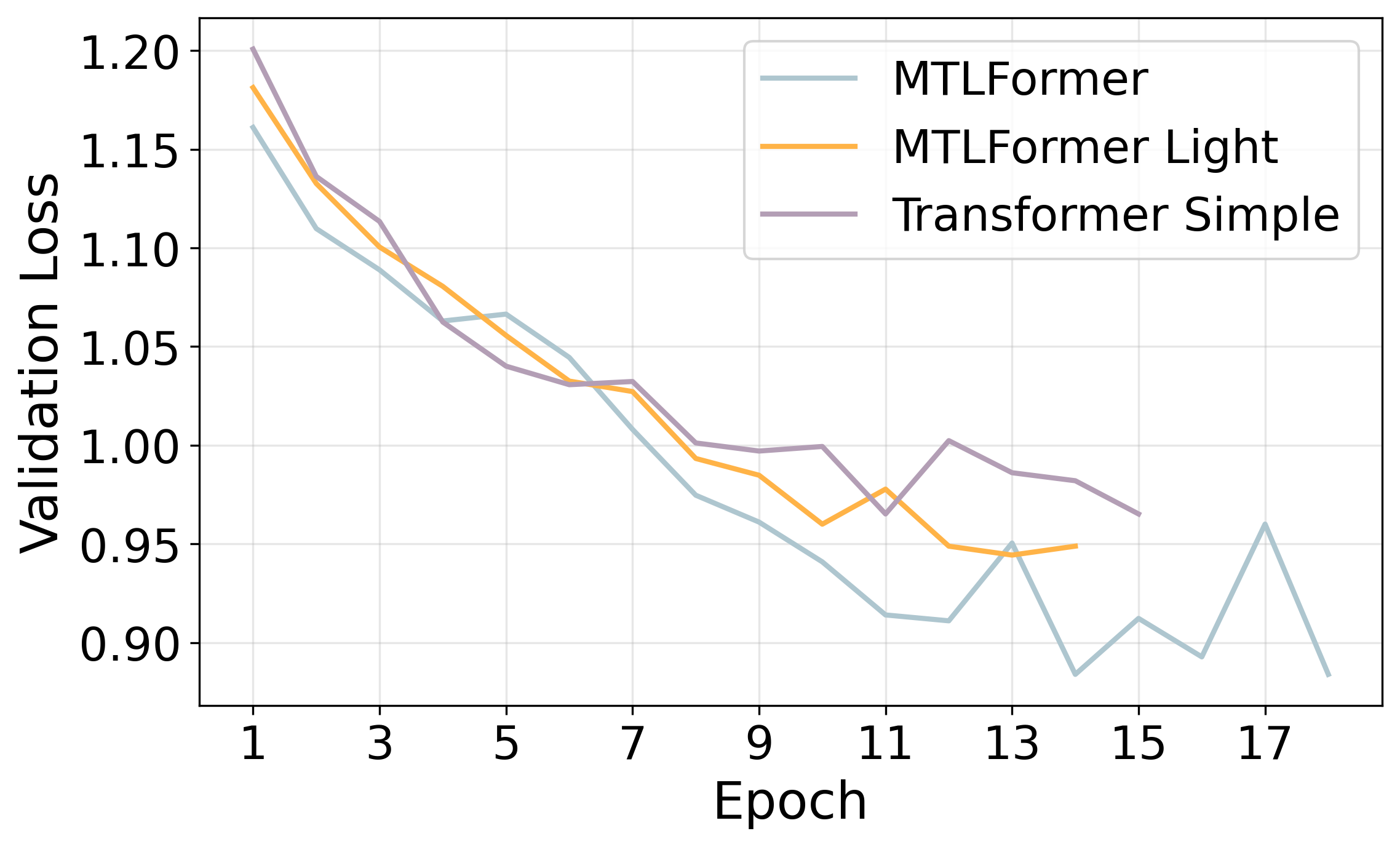}
    \caption{Dataset: BPIC2012W}
    \label{fig:val_loss_BPIC2012W}
  \end{subfigure}
  \hfill
  \begin{subfigure}[b]{0.45\textwidth}
    \includegraphics[width=\linewidth]{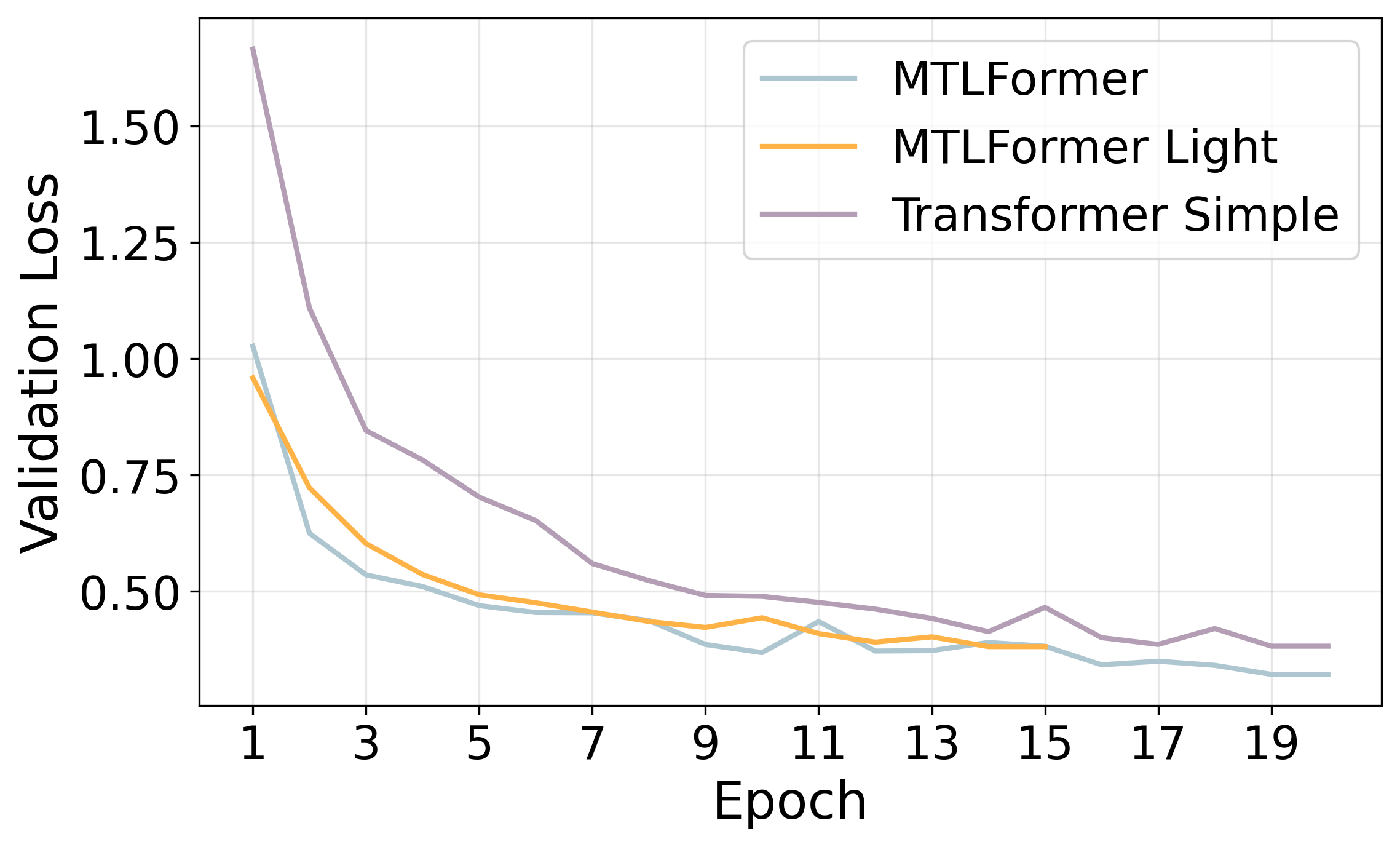}
    \caption{Dataset: P2P}
    \label{fig:val_loss_P2P}
  \end{subfigure}
  \caption{Validation loss curves for the three Transformer models.}
  \label{fig:val_loss_comparison}
\end{figure}

Furthermore, we analyze how the differently sized models compare in terms of validation loss progression.
As shown in \autoref{fig:val_loss_comparison}, the validation-loss curves for BPIC2012W and P2P reveal that the simplified variants converge at a similar pace than their full-sized counterparts. Remarkably, despite a significantly reduced parameter count, MTLFormer\textsubscript{light} often reaches its minimum validation loss in fewer epochs than the original MTLFormer. This suggests that model compactness does not hinder, and may even enhance, the optimization process in these settings. We observed similar convergence behavior across all other event logs, underscoring the robustness of the lightweight architectures.

% ... later in your document where you want the figures:

%% file: Tables/datasets.tex
\begin{table*}[t]
\centering
\setlength\tabcolsep{3pt} % default value: 6pt
\caption{Characteristics of the event logs.}
\label{tab:datasets}
\begin{tabular}{lrrrrr}
\toprule
\textbf{Event Log} & \textbf{Type} & \textbf{\#Traces} & \textbf{\#Events} & \textbf{\#Activities} & \textbf{\#Resources} \\ 
\midrule
 Production & real  & 225 & 4503 & 24 & 41 \\ 
 BPIC2012W & real  & 8616 & 59302 & 6 & 52 \\ 
 P2P & syn  & 608 & 9119 & 21 & 27 \\ 
 Confidential 1000 & syn  & 1000 & 38160 & 42 & 14 \\ 
 Confidential 2000 & syn  & 2000 & 77418 & 42 & 14 \\ 
 \bottomrule
\end{tabular}
\end{table*}

%% file: Tables/Transformer_results_LK.tex
\begin{table}[t!]
    \centering
    \setlength{\tabcolsep}{4pt}
    \caption{Results for the Transformer models for next activity prediction (NAP), next role prediction (NRP), next wait time prediction (NWTP), next duration prediction (NDP), and remaining time prediction (RTP).}
    \begin{tabular}{ccrrrrrr}
        \toprule
        Log & Model & NAP $\uparrow$ & NRP $\uparrow$ & NWTP $\downarrow$ & NDP $\downarrow$ & RTP $\downarrow$ & Parameters \\
        \midrule
        \multirow{4}{*}{\rotatebox{90}{P2P}} & MTLFormer & 0.84  & 0.93 & 3.32  & 0.11  & 24.41 & 93087 \\
        \noalign{\vskip 1mm}
        \cdashline{2-8}
        \noalign{\vskip 1mm}
        & MTLFormer$_\text{light}$ & 0.84  & 0.94  & 3.74  & 0.12  & 25.22 & 9519  \\
        & Transformer$_\text{simple}$ & 0.83  & 0.92  & 3.68  & 0.12 & 24.87 & 8407  \\

        \midrule
        \multirow{4}{*}{\rotatebox{90}{BPI12W}} & MTLFormer & 0.73  & 0.41 & 2.24  & 0.02  & 17.27 & 133 042 \\
        \noalign{\vskip 1mm}
        \cdashline{2-8}
        \noalign{\vskip 1mm}
        & MTLFormer$_\text{light}$ & 0.74  & 0.40 & 2.25  & 0.01  & 17.34 & 12514 \\
        & Transformer$_\text{simple}$ & 0.67  & 0.36 & 2.28  & 0.01  & 17.33 & 7954 \\

        \midrule
        \multirow{4}{*}{\rotatebox{90}{Prod.}} & MTLFormer &   0.21 & 0.33 & 1.53 & 0.2  & 37.03 & 169 003 \\
        \noalign{\vskip 1mm}
        \cdashline{2-8}
        \noalign{\vskip 1mm}
        & MTLFormer$_\text{light}$ & 0.15 & 0.29 & 1.56 & 0.20 & 36.66 & 26 555 \\
        & Transformer$_\text{simple}$ & 0.14 & 0.30 & 1.61 & 0.20 & 37.87 & 9823 \\

        \midrule
        \multirow{4}{*}{\rotatebox{90}{C.1000}} & MTLFormer  & 0.86 & 0.93  & 0.04  & 0.04 & 0.98 & 151 847 \\
        \noalign{\vskip 1mm}
        \cdashline{2-8}
        \noalign{\vskip 1mm}
        & MTLFormer$_\text{light}$ & 0.84 & 0.90 &  0.04 &  0.04 & 0.99 & 27 175 \\
        & Transformer$_\text{simple}$ & 0.81 & 0.91 & 0.04 & 0.04 & 1.00 & 28 163 \\

        \midrule
        \multirow{4}{*}{\rotatebox{90}{C.2000}} & MTLFormer & 0.86 & 0.93 & 0.03 & 0.03 & 0.92 & 135079 \\
        \noalign{\vskip 1mm}
        \cdashline{2-8}
        \noalign{\vskip 1mm}
        & MTLFormer$_\text{light}$ & 0.86 & 0.92 & 0.03 & 0.04 & 0.94 & 23 351 \\
        & Transformer$_\text{simple}$ & 0.85 & 0.92 & 0.03 & 0.04 & 0.94 & 46 975 \\

        \midrule
        \midrule
        \multirow{4}{*}{\rotatebox{90}{Average}} & MTLFormer & 0.70 & 0.71 & 1.43 & 0.08 & 16.12 & 136 412 \\
        \noalign{\vskip 1mm}
        \cdashline{2-8}
        \noalign{\vskip 1mm}
        & MTLFormer$_\text{light}$ & 0.69 & 0.69 & 1.52 & 0.08  & 16.23 & 19 823 \\
        & Transformer$_\text{simple}$ & 0.66  & 0.68 & 1.53 & 0.08 & 16.40 & 20 264  \\

        \bottomrule    
    \end{tabular}
    \label{tab:results_transformer}
\end{table}

%% file: Tables/LSTM_results_LK.tex
\begin{table}[t!]
    \centering
    \setlength{\tabcolsep}{4pt}
    \caption{Results for the LSTM models for next activity prediction (NAP), next role prediction (NRP), next wait time prediction (NWTP), next duration prediction (NDP), and remaining time prediction (RTP).}
    \begin{tabular}{ccrrrrrr}
        \toprule
        Log & Model & NAP $\uparrow$ & NRP $\uparrow$ & NWTP $\downarrow$ & NDP $\downarrow$ & RTP $\downarrow$ & Parameters \\
        \midrule
        \multirow{3}{*}{\rotatebox{90}{P2P}} & LSTM & 0.83  & 0.92 & 2.99 & 0.14 & 34.86 & 75 706\\
        \noalign{\vskip 1mm}
        \cdashline{2-8}
        \noalign{\vskip 1mm}
        & LSTM$_\text{light}$ & 0.81 & 0.88 & 3.55 & 0.16 & 36.16 & 20431 \\

        \midrule
        \multirow{3}{*}{\rotatebox{90}{BPI12}} & LSTM & 0.71 & 0.39 & 1.99 & 0.02 & 23.04 & 74 574 \\
        \noalign{\vskip 1mm}
        \cdashline{2-8}
        \noalign{\vskip 1mm}
        & LSTM$_\text{light}$ & 0.68 & 0.38 & 2.09 & 0.02 & 23.48 & 19 824 \\

        \midrule
        \multirow{3}{*}{\rotatebox{90}{Prod.}} & LSTM & 0.25 & 0.36 & 1.89 & 0.2 & 40.8 & 76 808 \\
        \noalign{\vskip 1mm}
        \cdashline{2-8}
        \noalign{\vskip 1mm}
        & LSTM$_\text{light}$ & 0.20 & 0.31 & 2.1 & 0.2 & 42.02 & 4368 \\

        \midrule
        \multirow{2}{*}{\rotatebox{90}{C.1000}} & LSTM & 0.87 & 0.92 & 0.03 & 0.04 & 1.16 & 76 146 \\
        \noalign{\vskip 1mm}
        \cdashline{2-8}
        \noalign{\vskip 1mm}
        & LSTM$_\text{light}$ & 0.86 & 0.92 & 0.03 & 0.05 & 1.2 & 20 671 \\

        \midrule
        \multirow{2}{*}{\rotatebox{90}{C.2000}} & LSTM & 0.86 & 0.92 & 0.02 & 0.04 & 1.1 & 76 146 \\
        \noalign{\vskip 1mm}
        \cdashline{2-8}
        \noalign{\vskip 1mm}
        & LSTM$_\text{light}$ & 0.86 & 0.91 & 0.03 & 0.04 & 1.15 & 20 671 \\

        \midrule
        \midrule
        \multirow{3}{*}{\rotatebox{90}{Avg.}} & LSTM & 0.70 & 0.70 & 1.38 & 0.09 & 20.19 & 75 876\\
        \noalign{\vskip 1mm}
        \cdashline{2-8}
        \noalign{\vskip 1mm}
        & LSTM$_\text{light}$ & 0.68 & 0.68 & 1.56 & 0.09 & 20.80 & 17 193 \\

        \bottomrule    
    \end{tabular}
    \label{tab:results_lstm}
\end{table}

%% file: Sections/05_Conclusion.tex
%Start 05_Conclusion.tex
\section{Conclusion}
\label{sec:conclusion}

In this work, we have investigated the feasibility of simplifying state‑of‑the‑art deep learning architectures for Predictive Process Monitoring (PPM), aiming to reduce computational overhead and, at the same time, preserve predictive performance.
In summary, our experiments across five diverse event logs show that deep PPM models can be significantly compressed without a proportional loss in predictive performance. Pruning both LSTM- and Transformer-based architectures, we found Transformers to be particularly robust: MTLFormer\textsubscript{light}, which preserves the architecture of its larger variant while reducing parameters by 85\%, retains 98–99\% of the original F\textsubscript{1} performance and shows only minimal increases in MAE for time predictions. 
By reducing the architecture to a single attention encoder, Transformer\textsubscript{simple} still retains 94–96\% of the original F\textsubscript{1} performance, demonstrating that even this streamlined design can compete with more complex models in PPM tasks. Moreover, MTLFormer\textsubscript{light} often converges more quickly during training than its larger counterpart, suggesting that compactness may facilitate the optimization process in certain cases.

In contrast, pruning the LSTM incurs more significant performance penalties, highlighting a fundamental difference in how these architectures leverage parameter capacity. These insights underscore the potential of lightweight attention models as the backbone for PPM systems. 

Future work could explore architecture-aware pruning techniques or neural architecture search to further optimize the trade-off between efficiency and accuracy. Most PPM tasks in this study are limited to next-event prediction; exploring more longer-term targets, such as suffix or outcome prediction, could provide valuable insights. 
Additionally, investigating the generalizability of lightweight models across unseen domains and their integration into real-time PPM systems offers promising directions for practical deployment.